\def\year{2022}\relax
\documentclass[letterpaper]{article} 
\usepackage{aaai22}  
\usepackage{times}  
\usepackage{helvet}  
\usepackage{courier}  
\usepackage[hyphens]{url}  
\usepackage{graphicx} 
\usepackage{natbib}  
\usepackage{caption} 
\DeclareCaptionStyle{ruled}{labelfont=normalfont,labelsep=colon,strut=off} 
\usepackage{algorithm}
\usepackage{algorithmic}

\usepackage{amsmath,amssymb,amsfonts}
\usepackage{booktabs}
\usepackage{subfigure}
\usepackage{tabularx}

\usepackage{newfloat}
\usepackage{listings}
\floatstyle{ruled}
\newfloat{listing}{tb}{lst}{}
\floatname{listing}{Listing}
\title{ShuttleNet: Position-aware Fusion of Rally Progress and Player Styles for Stroke Forecasting in Badminton}
\author{
    Wei-Yao Wang,
    Hong-Han Shuai,
    Kai-Shiang Chang,
    Wen-Chih Peng
}
\affiliations{

    National Yang Ming Chiao Tung University, Hsinchu, Taiwan \\
    \{sf1638.cs05, hhshuai, kevin5260523.cs05\}@nctu.edu.tw, wcpeng@cs.nctu.edu.tw
}

\begin{document}

\maketitle

\begin{abstract}
The increasing demand for analyzing the insights in sports has stimulated a line of productive studies from a variety of perspectives, \textit{e.g.}, health state monitoring, outcome prediction. In this paper, we focus on objectively judging what and where to return strokes, which is still unexplored in turn-based sports. By formulating stroke forecasting as a sequence prediction task, existing works can tackle the problem but fail to model information based on the characteristics of badminton. To address these limitations, we propose a novel Position-aware Fusion of Rally Progress and Player Styles framework (ShuttleNet) that incorporates rally progress and information of the players by two modified encoder-decoder extractors. Moreover, we design a fusion network to integrate rally contexts and contexts of the players by conditioning on information dependency and different positions. Extensive experiments on the badminton dataset demonstrate that ShuttleNet significantly outperforms the state-of-the-art methods and also empirically validates the feasibility of each component in ShuttleNet. On top of that, we provide an analysis scenario for the stroke forecasting problem.
\end{abstract}

\section{Introduction}
\label{sec:introduction}

In recent years, sports analytics has drawn significant attention due to the enormous market, which focuses on collecting sports data and implementing advanced techniques for mining useful information from the data.
In Major League Baseball, for example, teams started to shift to defense by moving infielders to specific positions according to the hitting pattern of opposing batters, and these types of shifts dramatically rose from 4.62\% in 2012 to 21.17\% in 2019 \cite{StatsPerform_baseball}.
Furthermore, there are about 100 sports-related organizations currently investigating new technologies for delivering interesting stream contents to fans in 2021 \cite{StatsPerform_fan}.
Generally, the target audience of sports analytics is composed of both coaching-oriented groups and community-oriented groups.
Coaching-oriented groups aim at improving player performance, \textit{e.g.}, tactic investigation \cite{DBLP:conf/kdd/DecroosHD18,DBLP:conf/atal/BealCNR20} and action valuing \cite{jayanth2018team,DBLP:conf/kdd/SiciliaPG19}, while community-oriented groups try to boost the spectator engagement, \textit{e.g.}, highlight prediction \cite{DBLP:conf/aaai/DecroosDHD17}, play retrieval \cite{DBLP:conf/kdd/WangLCJ19} and autonomous broadcast production \cite{Giancola_2021_CVPR}.
There are also sports analytics applications that are both for coaching-oriented groups and community-oriented groups~\cite{DBLP:conf/kdd/DecroosBHD19,DBLP:journals/corr/abs-2106-01786}, \textit{e.g.}, player performance analysis and providing the relation between performance and market value of the player.

\begin{figure}
  \centering
  \includegraphics[width=0.92\linewidth]{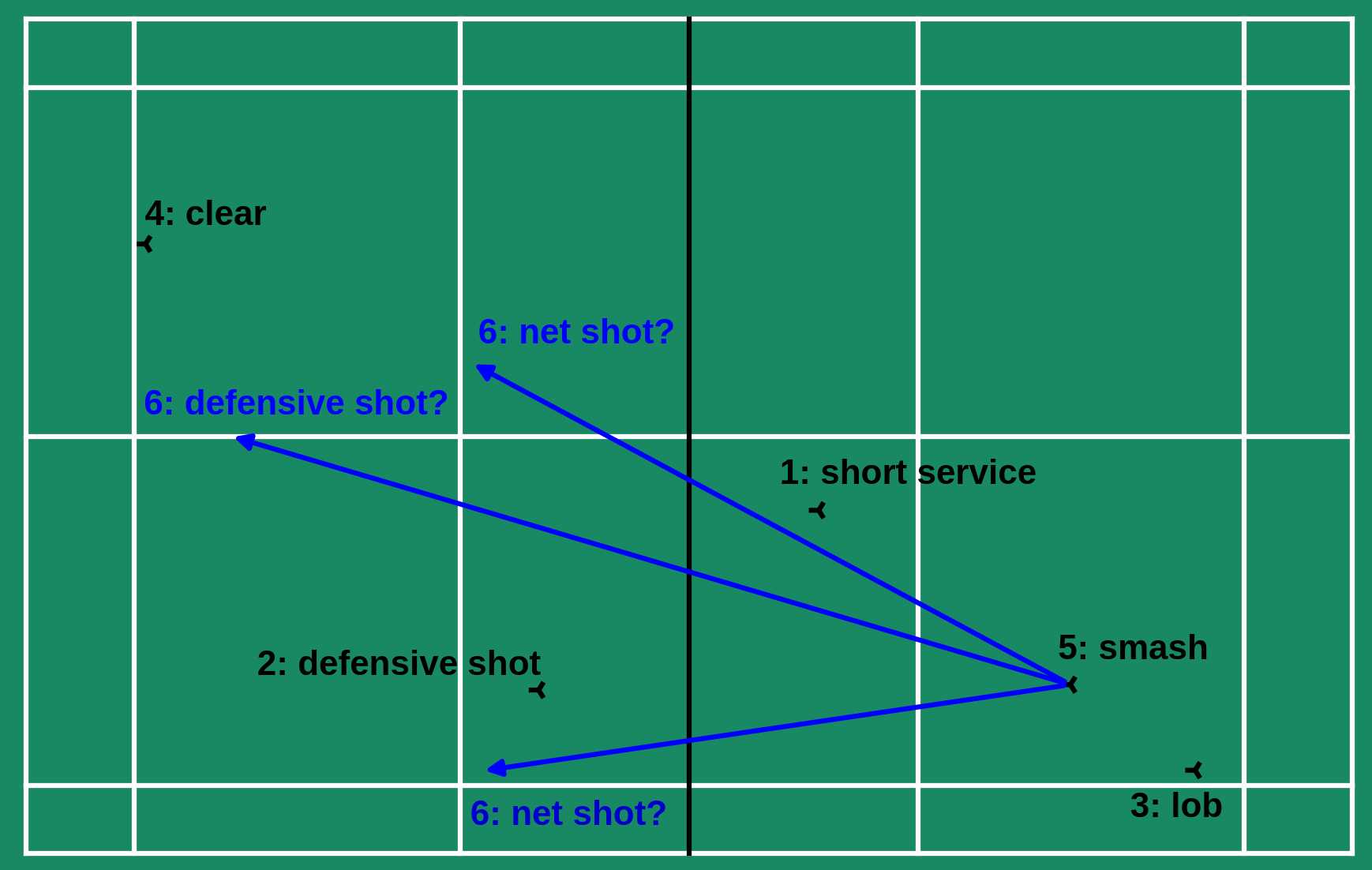}
  \caption{An example of stroke forecasting in a singles rally.
  The black line in the middle is the net.
  Each stroke consists of the order in the rally and its shot type.
  Blue lines with shot types represent possible choices in the next stroke.
  }
  \label{fig:example}
\end{figure}

In this paper, we focus on turn-based sports and use badminton as the demonstration example. The related works on badminton mainly focus on quantifying stroke performance \cite{DBLP:journals/jaihc/SharmaMKK21,DBLP:journals/corr/abs-2109-06431} or detecting stroke-related information from videos \cite{DBLP:conf/mir/ChuS17,DBLP:conf/apnoms/HsuWLCJIPWTHC19,Wang2020badminton,Yoshikawa2021}. 
However, there is another application that has not yet been addressed by previous works: to forecast the future strokes including shot types and locations given the past stroke sequences. Predicting future strokes based on the past strokes is essential and beneficial for coaching the player and determining the strategies since it simulates the tactics of the players, which can be used to investigate what shot types are often returned and where the strokes are returned to by the player for decision-making. In addition, stroke forecasting can also benefit the community for storytelling by assessing returning probability distributions during the matches.

Figure \ref{fig:example} illustrates an example of stroke forecasting.
Suppose the past five strokes with corresponding shot types and destination locations are known, and the fifth stroke is a smash from the left side, the player on the right side has several choices to return such as defensively returning to the middle of the left side, or returning close to the net to mobilize the opponent from the back court to front court. To the best of our knowledge, there is no existing method that can predict the next strokes.

To tackle this challenging problem, stroke forecasting can be formulated as the sequence prediction task. One possible solution is to use statistical methods like n-gram models in natural language processing to calculate the probabilities of occurrence for predicting next future strokes. Nevertheless, the probabilities of occurrence in n-gram models become sparse when increasing window size. To solve the issue, the sequence-to-sequence model \cite{DBLP:conf/nips/SutskeverVL14}  can be applied to encode the input sequence and then decode the output sequence with encoding vectors. However, there are three challenges for applying the sequence-to-sequence model directly for stroke forecasting. 1) \textit{Mixed sequence}. One of the characteristics of badminton is that there are two players returning strokes alternatively to form a rally.
Therefore, stroke forecasting is a turn-based sequence prediction task rather than a conventional sequence prediction task with the same target in the sequence. 2) \textit{Multiple outputs.} Differing from general sequence tasks that only predict one output, the stroke forecasting task has multiple outputs (shot types and area coordinates) at each timestamp.
3) \textit{Player dependence}. Returning strokes are based on the overall styles of the players and the current situation in the rally.
Furthermore, the importance of the overall styles of the players and the current situation in the rally also varies when encountering different players and different positions. It is challenging to disentangle the player features directly from the rally sequences.



To address the aforementioned challenges, we propose a novel Position-aware Fusion of Rally Progress and Player Styles framework (ShuttleNet), which consists of two encoder-decoder extractors for modeling rally progress and retrieving player styles from turn-based sequence and a fusion network to take into account the dependencies between rally progress and player styles at each stroke.
To predict multiple outputs at each step, two task-specific predictors are adopted in the end for predicting shot types and area coordinates. Specifically, the first encoder-decoder extractor, named Transformer-Based Rally Extractor (TRE), is designed to capture the progress of the rally.
Moreover, the second encoder-decoder extractor, named Transformer-Based Player Extractor (TPE), separates the information of each player to generate the context of each player.
Finally, a Position-aware Gated Fusion Network (PGFN) is adopted to fuse rally contexts and contexts of two players by incorporating information weights and position weights. In this manner, we can learn different contributions at each stroke to predict future shot types and area coordinates.
In summary, our contributions are as follows:
\begin{itemize}
    \item A novel framework named Position-aware Fusion of Rally Progress and Player Styles (ShuttleNet) is proposed to predict future strokes by giving past observed strokes.
    To the best of our knowledge, this is the first work for stroke forecasting in sports, which can be applied to turn-based sports analytics.
    \item The proposed framework first generates rally contexts and contexts of players by leveraging two encoder-decoder extractors and then fuses these contexts based on information weights and position weights.
    Furthermore, we introduce an attention mechanism to better integrate the information of shot types and locations.
    \item Extensive experiments and ablation studies on a real-world badminton dataset are conducted to demonstrate the effectiveness of the proposed ShuttleNet framework.
\end{itemize}

\section{Related Works}
\label{sec:related-work}

\subsection{Sport data analytics}
The area of artificial intelligence for sports contains five technical issues \cite{DBLP:conf/ijcai/DecroosBHD20}: representation~\cite{DBLP:conf/kdd/DecroosHD18}, interpretability~\cite{silverbaseball}, decision making~\cite{DBLP:conf/kdd/SiciliaPG19}, understanding behavior~\cite{DBLP:conf/cvpr/WeeratungaDH17}, and experimental evaluation \cite{DBLP:conf/ijcai/LiuS18}. For instance, \citet{DBLP:conf/kdd/DecroosHD18} proposed SPADL to address the data science challenges as unique definitions of different vendors by unifying soccer event-based data, which reduces the burden on redesigning data formats when serving different objectives.
Mimic learning is another approach used in sports that builds a model with both accurate predictions and interpretable insights \cite{DBLP:conf/kdd/SunDSL20}.
Action valuing designs objective metrics by valuing scoring and defensive performance of each action, which can be used as evaluation tools for understanding the behavior of players \cite{DBLP:conf/kdd/DecroosBHD19,DBLP:journals/corr/abs-2106-01786}.
Outcome prediction utilizes machine learning approaches and has been applied to cricket, soccer, and badminton, to help coaches to select the optimal players to win the game \cite{jayanth2018team,DBLP:conf/kdd/RobberechtsHD21,DBLP:journals/jaihc/SharmaMKK21}.
\citet{DBLP:journals/tist/PappalardoCFMPG19} proposed PlayeRank by combining multi-dimensional and role-aware evaluations from a massive soccer database, in order to provide evaluations and rankings of soccer players.
Our focus, in contrast, addresses stroke forecasting in badminton, which is also critically related to the above issues.
Predicting future strokes can facilitate decision-making and provide behavioral understanding of returning strokes within probability distributions.
Moreover, ground truth labels can also be objectively obtained when labeling instead of customizing evaluation methods.

\begin{figure*}
  \centering
  \includegraphics[width=0.98\linewidth]{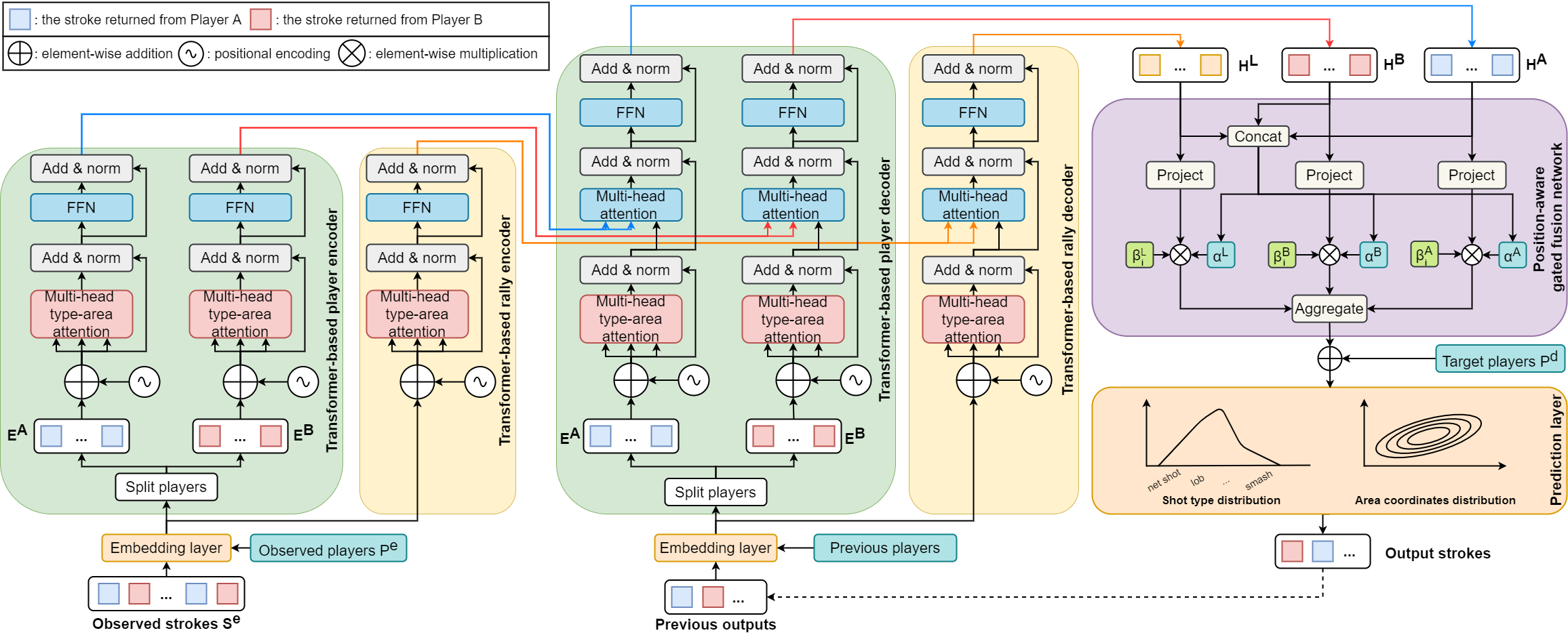}
  \caption{Illustration of the ShuttleNet framework.
  The Transformer-based rally extractor (TRE) in the yellow part generates rally contexts.
  The Transformer-based player extractor (TPE) in green shares the same set of parameters and generates contexts of both players.
  The contexts of the rally and both players are fed into the position-aware gated fusion network (PGFN) to weigh the contributions based on information and position for predicting future strokes.}
  \label{fig:framework-overview}
\end{figure*}

\subsection{Sequence Prediction}
The sequence-to-sequence model \cite{DBLP:conf/nips/SutskeverVL14} was proposed to deal with machine translation by encoding input with a Long Short-Term Memory (LSTM) \cite{DBLP:journals/neco/HochreiterS97} and then using another LSTM as a decoder to generate predictions.
The encoder-decoder architecture has been widely adopted for various sequence prediction tasks such as pedestrian trajectory prediction \cite{DBLP:conf/cvpr/MohajerinR19} and citation forecasting \cite{DBLP:conf/aaai/LiuYXWZC17}.
\citet{DBLP:conf/aaai/XuYD20} designed CF-LSTM with a feature-cascaded module to robustly capture dynamic information of trajectories from previous steps without other pedestrians' information. More recently, Transformer has become popular for sequence prediction. \citet{DBLP:conf/icpr/GiuliariHCG20}, for example, adopted Transformer Networks to predict future trajectory.
DMA-Nets introduced two temporal attention mechanisms to model local temporal information and global temporal information for citation forecasting \cite{DBLP:conf/aaai/JiSFCRL21}, which is the most similar setting (multiple outputs) to stroke forecasting. However, these previous works focus on the same target in a sequence, and thus cannot be directly applied to turn-based sequences.

\section{Problem Formulation}
\label{sec:problem}

Let $R=\{S_r, P_r\}_{r=1}^{|R|}$ denote historical rallies of badminton matches, where the $r$-th rally is composed of a stroke sequence with type-area pairs $S_r=(\langle s_1, a_1\rangle,\cdots,\langle s_{|S_r|}, a_{|S_r|}\rangle)$ and a player sequence $P_r=(p_1,\cdots,p_{|S_r|})$.
At the $i$-th stroke, $s_i$ represents the shot type, $a_i=\langle x_i, y_i\rangle \in \mathbb{R}^{2}$ are the coordinates of the shuttle destinations, and $p_i$ is the player who hits the shuttle. We denote Player A as the served player and Player B as the other for each rally in this paper. For instance, given a singles rally between Player A and Player B, $P_r$ may become $(A, B, \cdots, A, B)$.
We formulate the problem of stroke forecasting as follows. For each rally, given the observed $\tau$ strokes $(\langle s_i, a_i\rangle)_{i=1}^{\tau}$ with players $(p_i)_{i=1}^{\tau}$, the goal is to predict the future strokes including shot types and area coordinates for the next $n$ steps, i.e., $(\langle s_i, a_i\rangle)_{i={\tau+1}}^{\tau+n}$.

\section{Methodology}
\label{sec:methodology}

Figure \ref{fig:framework-overview} illustrates the overview of the proposed framework.
The input of the encoder side is the sequence of observed $\tau$ strokes $S^{e}=(\langle s_i, a_i\rangle)_{i=1}^{\tau}$ and players $P^{e}=(p_i)_{i=1}^{\tau}$, and the decoder auto-regressively predicts the sequence of the future $n$ strokes $S^{d}=(\langle s_i, a_i\rangle)_{i={\tau+1}}^{\tau+n}$ by taking encoding contexts and target players $P^{d}=(p_i)_{i={\tau+1}}^{\tau+n}$.
Each stroke encompassing a shot type and area coordinates is embedded with player information from the embedding layer as a personal stroke.
Each encoder-decoder extractor is based on the Transformer~\cite{DBLP:conf/nips/VaswaniSPUJGKP17}. We replace the first multi-head self-attention layer in both the encoder and decoder with the proposed type-area-attention layer to better integrate the information of shot types and area. Moreover, contexts of the rally are generated by the Transformer-based rally extractor, and contexts of the two players are obtained by the Transformer-based player extractor.
Outputs of these contexts are fused by the position-aware gated fusion network using information weights and position weights for predicting future shot types and area coordinates.

\subsection{Embedding Layer}
Each stroke contains a shot type and area coordinates with the player who hits the stroke. The output of embedding layer at $i$-th stroke $e_i$ is calculated as follows:
\begin{equation}
  e_i = \langle e_i^s, e_i^a\rangle = \langle s_i' + p_i', a_i' + p_i'\rangle,
  \label{eq:embedding_layer}
\end{equation}
where $s_i'$ is a type embedding projected from $s_i$ using $M^s \in \mathbb{R}^{N_s \times d}$, where $N_s$ is the number of shot types, $p_i'$ is a player embedding projected from $p_i$ using $M^p \in \mathbb{R}^{N_p \times d}$, where $N_p$ is the number of players, and $a_i'$ is an area embedding projected from $a_i$ using $M^a \in \mathbb{R}^{2 \times d}$ with the ReLU activation function.
In order to make use of the player in shot types and area, player embeddings are added to both type embeddings and area embeddings.
The parameters of embedding layers in the encoder side and the decoder side are shared similar to \cite{DBLP:conf/eacl/PressW17} for size reduction.

\subsection{Transformer-based Rally Extractor (TRE)}
TRE reflects the current situation in the rally, which is a critical condition for returning strokes.
For example, if the player defensively returns a stroke like a lob to the back court, this indicates the player may become passive and the other player can seize the chance to return an aggressive stroke.

To capture the progress in the rally, we first add positional encodings \cite{DBLP:conf/nips/VaswaniSPUJGKP17} to the embeddings by
\begin{equation}
\begin{aligned}
  E^L &= (\langle \tilde{e}^s_1, \tilde{e}^a_1\rangle, \langle \tilde{e}^s_2, \tilde{e}^a_2\rangle, \cdots)\\
      &= (\langle e_1^s+pe_1, e_1^a+pe_1\rangle, \langle e_2^s+pe_2, e_2^a+pe_2\rangle, \cdots),
  \label{rally_position}
\end{aligned}
\end{equation}
where $pe_i$ is the position encoding for $i$-th stroke.

Afterward, we adopt a modified Transformer framework by replacing the first multi-head self-attention layer in the encoder and decoder with the proposed multi-head type-area-attention layer. Specifically, we take $E^L$ as the inputs of the Transformer-based rally extractor and generate the contexts of the rally $H^L=(h^L_{\tau+1}, h^L_{\tau+2},\cdots)$, where the $i$-th stroke $h^L_i \in \mathbb{R}^{d}$ is a $d$ dimension vector.

\subsubsection{Type-area-attention layer}
Since there are two components (shot type and area) of each stroke, the self-attention mechanism can only be applied in an early-fusion manner (\textit{e.g.}, concatenation), which forces to attend shot types and area at the same position. However, it is expected that playing strategies should be considered from different aspects in badminton matches.
For example, returning a current shot type may be considered the last shot type to decide the proper choice, while where to return may be considered the previous location returned from the same player because the opponent is weak on a specific side.

Therefore, inspired by disentangled attention \cite{DBLP:conf/iclr/HeLGC21}, we propose an attention mechanism to separately characterize the importance of shot types and area and then aggregate corresponding scores as final attention scores. Here, we illustrate the computation of attention contexts on the encoder side, and the decoder follows a similar process. Given the input sequence with positional type embeddings $E^s=(\tilde{e}_1^s,\cdots,\tilde{e}_{\tau}^s)$ and positional area embeddings $E^a=(\tilde{e}_1^a,\cdots,\tilde{e}_{\tau}^a)$, the formula of the multi-head type-area attention is derived as follows:
\begin{equation}
  Q_s=E^sW^{Q_s},K_s=E^sW^{K_s},V_s=E^sW^{V_s},
  \label{shot_qkv}
\end{equation}
\begin{equation}
  Q_a=E^aW^{Q_a},K_a=E^aW^{K_a},V_a=E^aW^{V_a},
  \label{area_qkv}
\end{equation}
\begin{equation}
  A = Q_aK_a^{T} + Q_aK_s^{T} + Q_sK_a^{T} + Q_sK_s^{T},
  \label{attention_score}
\end{equation}
\begin{equation}
  TAA(E_s, E_a) = softmax(\frac{A}{\sqrt{4d}})(V_a+V_s),
  \label{attention_output}
\end{equation}
\begin{equation}
  MultiHead(E_s, E_a) = Concat(TAA_1,\cdots,TAA_h)W^o,
  \label{multi_head}
\end{equation}
where $TAA$ denotes the function of type-area attention with single head, $Q_s$, $K_s$, and $V_s$ are queries, keys and values of $E^s$ projected using projection matrices $W^{Q_s}, W^{K_s}, W^{V_s} \in \mathbb{R}^{d \times d}$, respectively.
$Q_a$, $K_a$, and $V_a$ are queries, keys and values of $E^a$ projected using matrices $W^{Q_a}, W^{K_a}, W^{V_a} \in \mathbb{R}^{d \times d}$, respectively.
$h$ is the number of heads, and $W^o \in \mathbb{R}^{hd \times d}$ is a learnable matrix.

\subsection{Transformer-based Player Extractor (TPE)}
In addition to the information of the rally, returning strokes also needs to consider the overall style of each player.
That is, the player should minimize their opponents' advantages and maximize their own.
To this end, we designed an extractor to split the sequence into two subsequences based on the player and then to produce the contexts of each player.

First, the outputs of the embedding layer are alternatively split based on the players as follows:
\begin{equation}
  E^A = (e_1, e_3, \cdots),
  E^B = (e_2, e_4, \cdots),
  \label{split_players}
\end{equation}
where $E^A$ is the sequence of Player A, $E^B$ is the sequence of Player B. TPE adopts two encoder-decoder architectures to capture the two sequences split by players, both of which are the same as the architecture in TRE.
Specifically, $E^A$ and $E^B$ are fed into TPE to generate corresponding contexts.

It is worth noting that the positional encodings are added separately to the two subsequences to specify the order of each player rather than using the entire sequence in Equation \ref{rally_position}.
Further, the parameters of the two architectures are shared not only to reduce the number of parameters but also to prevent player information from falling on the same side, which would cause imbalance.

Since the lengths of two subsequences are shorter than the original sequence, sequence alignment is applied to align two subsequences with the same length of rally sequence after generating two contexts of the players.
The alignment principle is to add a copy stroke to the next stroke, which becomes the opponent to return.
The formula of the sequence alignment is derived as follows:
\begin{equation}
  H^A = (h^A_{\tau+1}, h^A_{\tau+1}, h^A_{\tau+2}, h^A_{\tau+2},\cdots),
  \label{sequence_alignment_A}
\end{equation}
\begin{equation}
  H^B = (0, h^B_{\tau+1}, h^B_{\tau+1}, h^B_{\tau+2}, h^B_{\tau+2},\cdots),
  \label{sequence_alignment_B}
\end{equation}
where the $i$-th stroke $h^A_i \in \mathbb{R}^{d}$ denotes the output from the decoder for Player A, and the $i$-th stroke $h^B_i \in \mathbb{R}^{d}$ is the output from the other decoder for Player B.
Zero is padded at the first stroke of $H^B$ since the first stroke is always served by Player A.

\subsection{Position-aware Gated Fusion Network (PGFN)}
When returning strokes, players consider various important information about both players and current rally.
Moreover, the importance of these types of information at each stroke will vary. To take the above consideration into the design, we propose a position-aware gated fusion network based on gated multi-modal units \cite{DBLP:journals/nca/OvalleSMG20} to fuse rally contexts and contexts of two players.

Given the contexts of Player A and Player B ($h^A_i$ and $h^B_i$), and the rally $h^L_i$ at $i$-th stroke, the PGFN first projects to hidden vectors of fusing contexts:
\begin{equation}
  \tilde{h}_i^A = \delta_t(h^A_iW^A),
  \tilde{h}_i^B = \delta_t(h^B_iW^B),
  \tilde{h}_i^L = \delta_t(h^L_iW^L),
  \label{hidden}
\end{equation}
where $\delta_t(\cdot)$ is the tanh activation function, and $W^A, W^B, W^L \in \mathbb{R}^{d \times d}$ are learnable matrices.
The information weights to represent the importance of the three contexts are calculated as follows:
\begin{equation}
  \alpha^A = \delta_s( [\tilde{h}_i^A,\tilde{h}_i^B,\tilde{h}_i^L]\tilde{W}^A),
  \label{information_weights_A}
\end{equation}
\begin{equation}
  \alpha^B = \delta_s( [\tilde{h}_i^A,\tilde{h}_i^B,\tilde{h}_i^L]\tilde{W}^B),
  \label{information_weights_B}
\end{equation}
\begin{equation}
  \alpha^L = \delta_s( [\tilde{h}_i^A,\tilde{h}_i^B,\tilde{h}_i^L]\tilde{W}^L),
  \label{information_weights_L}
\end{equation}
where $\delta_s(\cdot)$ is the sigmoid activation function, $[\cdot,\cdot,\cdot]$ denotes the concatenation operator, and $\tilde{W}^A, \tilde{W}^B, \tilde{W}^L \in \mathbb{R}^{3d \times d}$ are learnable matrices.

Finally, the $i$-th fusing output is calculated as:
\begin{equation}
  z_i=\delta_s(\beta_i^A \otimes \alpha^A \otimes \tilde{h}_i^A+\beta_i^B \otimes \alpha^B \otimes \tilde{h}_i^B+\beta_i^L \otimes \alpha^L \otimes \tilde{h}_i^L),
  \label{fusion_output}
\end{equation}
where $\otimes$ denotes the element-wise multiplication, and $\beta_i^A, \beta_i^B, \beta_i^L \in \mathbb{R}^{d}$ are learnable position weights to learn how much to pass at each stroke.

\subsection{Prediction Layer}
To predict the shot type and area coordinates at $i$-th stroke, we first assume area coordinates follow a bi-variate Gaussian distribution since there exists uncertainty and potential multi-modality when returning strokes.
For instance, when the opponent returns the stroke to the back court, the player can return to the non-handedness side to force the opponent to return the shuttle with back hand, or to return to near the net to make the opponent return defensively.
Moreover, the predictive distribution enables the ability to investigate the locations of frequent and less frequent stroke returns to better understand the players' behaviors.
Specifically, area coordinates are sampled from a bi-variate Gaussian distribution with the mean $\mu_{i}=\langle \mu_x, \mu_y\rangle_{i}$, standard deviation $\sigma_{i}=\langle \sigma_x, \sigma_y\rangle_{i}$, and correlation coefficient $\rho_{i}$.

Hard parameter sharing is adopted to share the same fusion outputs to predict multiple outputs at each step.
Two linear layers are used to predict the parameterized distribution $\langle \mu_{i+1},\sigma_{i+1},\rho_{i+1}\rangle$ and the shot type $\hat{s}_{i+1}$ at $(i+1)$-th stroke by combining the target player embedding $p_{i+1}$ and the fusing output $z_{i}$, respectively:
\begin{equation}
  \hat{s}_{i+1}=softmax((z_i+p_{i+1})W^s),
  \label{predict_shot}
\end{equation}
\begin{equation}
  \langle \mu_{i+1},\sigma_{i+1},\rho_{i+1}\rangle=(z_i+p_{i+1})W^a,
  \label{predict_area}
\end{equation}
where $W^s \in \mathbb{R}^{d \times N_s}$ and $W^a \in \mathbb{R}^{d \times 5}$ are two learnable matrices.
The predicted area coordinates are sampled by $\langle \hat{x}_{i+1},\hat{y}_{i+1}\rangle \sim \mathcal{N} (\mu_{i+1},\sigma_{i+1},\rho_{i+1})$.
The reason for adding the target player embedding to the fused contexts is to specify the player who returns the stroke.

We minimize cross-entropy loss $\mathcal{L}_{type}$ to learn the prediction of shot types:
\begin{equation}
  \mathcal{L}_{type} = -\sum_{r=1}^{|R|}\sum_{i=\tau+1}^{|S_r|} s_i log(\hat{s}_i).
  \label{shot_loss}
\end{equation}
We also minimize the negative log-likelihood loss $\mathcal{L}_{area}$ to learn the prediction of area coordinates:
\begin{equation}
  \mathcal{L}_{area}=-\sum_{r=1}^{|R|}\sum_{i=\tau+1}^{|S_r|} log(\mathcal{P}(x_i, y_i | \mu_i,\sigma_i,\rho_i)).
  \label{area_loss}
\end{equation}
The total loss $\mathcal{L}$ of our model is jointly trained with:
\begin{equation}
  \mathcal{L}=\mathcal{L}_{type}+\mathcal{L}_{area}.
  \label{total_loss}
\end{equation}

\section{Results and Analysis}
\label{sec:experiment}

\subsection{Experimental Setup}
\noindent\textbf{Dataset.} Since there is no public dataset of stroke event records, we collected real-world badminton singles matches from public sources\footnote{http://bwf.tv} and asked domain experts to manually label them.
The dataset contains 75 high-ranking matches from 2018 to 2021 played by 31 players from men's singles and women's singles.
After filtering flaw data, \textit{e.g.}, replay highlights, the dataset contains 180 sets, 4,325 rallies, and 43,191 strokes.
The average length of the rallies is 10.
There are 10 shot types defined by domain experts for distinguishing the strokes:
\textit{
  net shot, 
  clear, 
  push/rush, 
  smash, 
  defensive shot, 
  drive, 
  lob, 
  drop, 
  short service, 
}and \textit{long service.} 

For the stroke forecasting task, each stroke contains the id of the rally, the order of the stroke in a rally, the player returning the stroke, the shot type, and the area coordinates where the shuttle was returned to. We split the first 80\% of the rallies of each match as training data to ensure that the model is equipped with past information of all players, and the remaining rallies were used for testing. We conducted 5-fold cross-validation for tuning hyper-parameters.

\begin{table*}
    \centering
    \begin{tabular}{c|ccc|ccc|cccccc}
    \toprule
    & \multicolumn{3}{c|}{$\tau=8$} & \multicolumn{3}{c|}{$\tau=4$} & \multicolumn{3}{c}{$\tau=2$} \\
    \cmidrule{2-10}
    Model & CE & MSE & MAE & CE & MSE & MAE & CE & MSE & MAE \\
    \midrule
        Seq2Seq                     & 2.5219 & 1.7124 & 1.4181 &
                                      2.5192 & 1.6674 & 1.4049 &
                                      2.5325 & 1.6799 & 1.4022 \\
        CF-LSTM                     & \underline{2.3138} & 2.1805 & 1.6844 &
                                      \underline{2.2623} & 2.2510 & 1.7055 &
                                      \underline{2.3860} & 2.0392 & 1.5966 \\
        TF                          & 2.3843 & \underline{1.6427} & \underline{1.4017} &
                                      2.3881 & 1.6665 & 1.4033 & 
                                      2.4243 & 1.6317 & 1.3773 \\
        dNRI                        & 2.4391 & 2.4056 & 1.7903 &
                                      2.4475 & 2.3518 & 1.7822 &
                                      2.4441 & 2.3025 & 1.7587 \\
        DMA-Nets                    & 2.4949 & 1.8419 & 1.4791 &
                                      2.6710 & 1.8463 & 1.4876 &
                                      2.5975 & 1.8436 & 1.4813 \\
    \midrule
        ShuttleNet\textsubscript{P2R} (Ours)  & 2.3892 & 1.6665 & 1.4052 &
                                             2.3112 & 1.6296 & \textbf{1.3838} &
                                             2.3963 & 1.5900 & \underline{1.3693} \\
        ShuttleNet\textsubscript{R2P} (Ours)  & 2.3528 & 1.6864 & 1.4233 & 
                                             2.3874 & \underline{1.6278} & \underline{1.3882} &
                                             2.3923 & \textbf{1.5627} & \textbf{1.3563} \\
    \midrule    
        ShuttleNet (Ours)                     & \textbf{1.9802} & \textbf{1.5856} & \textbf{1.3802} &
                                             \textbf{1.9916} & \textbf{1.5867} & 1.3896 &
                                             \textbf{2.0755} & \underline{1.5761} & 1.3747 \\
    \bottomrule
\end{tabular}
    \caption{Quantitative results of our models and baselines on different given lengths. The best result in each column is in boldface while the second best result is underlined.}
    \label{tab:experiment-performance}
\end{table*}

\noindent\textbf{Implementation Details.} The dimension of embeddings and contexts ($d$) was set to 32, the number of heads ($h$) used in multi-head attention and multi-head type-area attention was set to 2, and the inner dimension of feed-forward layer was 64.
The max sequence length of a rally was 35.
$n$ is the rally length and varies in different rallies.
The layer normalization \cite{DBLP:journals/corr/BaKH16} and dropout technique with a dropout rate of 0.1 were used for each sublayer, similar to \cite{DBLP:conf/nips/VaswaniSPUJGKP17}.
The batch size was 32 and the number of training epochs was 150 using Adam \cite{DBLP:journals/corr/KingmaB14} as the optimizer. The learning rate was set to 0.0001.
In the training phase, we adopted zero padding for sequences and used ground truth labels as the next step input in decoding. In the evaluation phase, we replaced the ground truth labels with sampled shot types and area coordinates.
Following the procedures for evaluating stochastic models in previous work, \textit{e.g.}, \cite{DBLP:conf/cvpr/AmirianHP19}, we generated $K=10$ samples and took the closest one to ground truth for evaluation.
The input of area coordinates was normalized with the mean as zero. All the training and evaluation phases were conducted on a machine with Intel i7-8700 3.2GHz CPU, Nvidia GTX 2070 8GB GPU, and 32GB RAM. The average results from 10 runs are reported.
Our code is available at https://github.com/wywyWang/ShuttleNet.

\subsubsection{Baselines}
Due to the lack of baselines for the proposed tasks, we compared the proposed model with the baselines of various sequence prediction tasks:
\begin{itemize}
    \item Seq2Seq \cite{DBLP:conf/nips/SutskeverVL14} consists of one LSTM as encoder and another LSTM as decoder.
    \item CF-LSTM \cite{DBLP:conf/aaai/XuYD20} is a feature-cascaded LSTM integrating feature information from previous two steps as dynamic interactions.
    \item TF \cite{DBLP:conf/icpr/GiuliariHCG20} utilizes the Transformer Network to learn contexts of pedestrians.
    \item dNRI \cite{Graber_2020_CVPR} models dynamic entity relations for neural relational inference.
    \item DMA-Nets \cite{DBLP:conf/aaai/JiSFCRL21} constructs a hierarchical dynamic attention layer by considering local temporal information and global temporal information for citation forecasting, which is the closest setting to stroke forecasting with multiple outputs at each timestamp.
    
\end{itemize}
Due to the lack of official codes, we reproduced CF-LSTM and DMA-Nets by following the corresponding implementation details in their papers.
For fair comparison, the same embedding layer, prediction layer, and hyper-parameters were used for all the baselines. Moreover, since all the baselines take single inputs\footnote{DMA-Nets has multiple inputs but also using concatenation.}, we concatenated shot types and area and projected them to same dimension of these baselines instead.
In addition, the loss function of shot types was added to the baselines to fit the stroke forecasting task.

To better explore the design of our proposed method, we extended two variants of our method in the experiments:
\begin{itemize}
    \item ShuttleNet\textsubscript{P2R} feeds the outputs of the embedding layer first to TPE and then feeds the outputs of TPE into TRE.
    \item ShuttleNet\textsubscript{R2P} feeds the outputs of the embedding layer first to TRE and then feeds the outputs of TRE into TPE.
\end{itemize}

\subsection{Quantitative Results}
\noindent\textbf{Comparison with Baselines.}
To evaluate the results of shot type prediction, we used cross-entropy (CE), which has been widely used for uncertainty measurement \cite{DBLP:conf/emnlp/SchmidtMH19}. Moreover, mean absolute error (MAE) and mean square error (MSE) were used for evaluating the predicted area coordinates similar to \cite{Graber_2020_CVPR}. We conducted three sets of experiments with the number of observed strokes $\tau$ set to 8, 4, and 2 to investigate the performance with different numbers of observed strokes. Table \ref{tab:experiment-performance} reports the best results of different models, which shows that our model consistently outperforms other baselines for both shot types and area prediction in terms of all metrics and different $\tau$. Specifically, our method surpasses all the baselines by at least 12.0\% and 3.4\% in terms of CE and MSE, respectively. 


As CF-LSTM and dNRI are based on the characteristics of trajectories (such as velocity), they fail to perform well on forecasting area coordinates. In other words, the trajectories in the dataset are different from the human trajectories since they are event-based data and dramatically change positions.
Moreover, Seq2Seq, TF, and DMA-Nets are biased in their ability to forecast area coordinates well, which indicates that integrating rally information is insufficient for returning strokes.
Also, these methods lack the capacity for turn-based targets in a sequence since they assume each element in the sequence belongs to the same target.
With the Transformer-based player extractor that considers player information, our model is capable of extracting the contexts of the players to correctly predict future strokes.

It is worth noting that the performance of both variants of ShuttleNet\textsubscript{P2R} and ShuttleNet\textsubscript{R2P} did not improve for shot types but improved for area coordinates. This indicates the effect on applying the fusing technique.
Early integration of player contexts when extracting the rally information and vice versa, both hamper the model's learning of the other information. Our model thus demonstrates the need of learning each type of information separately and then employing the fusing technique afterward to achieve the best results.

\subsubsection{Ablation Study}

\begin{figure*}[t]
  \centering
  \includegraphics[width=0.91\linewidth]{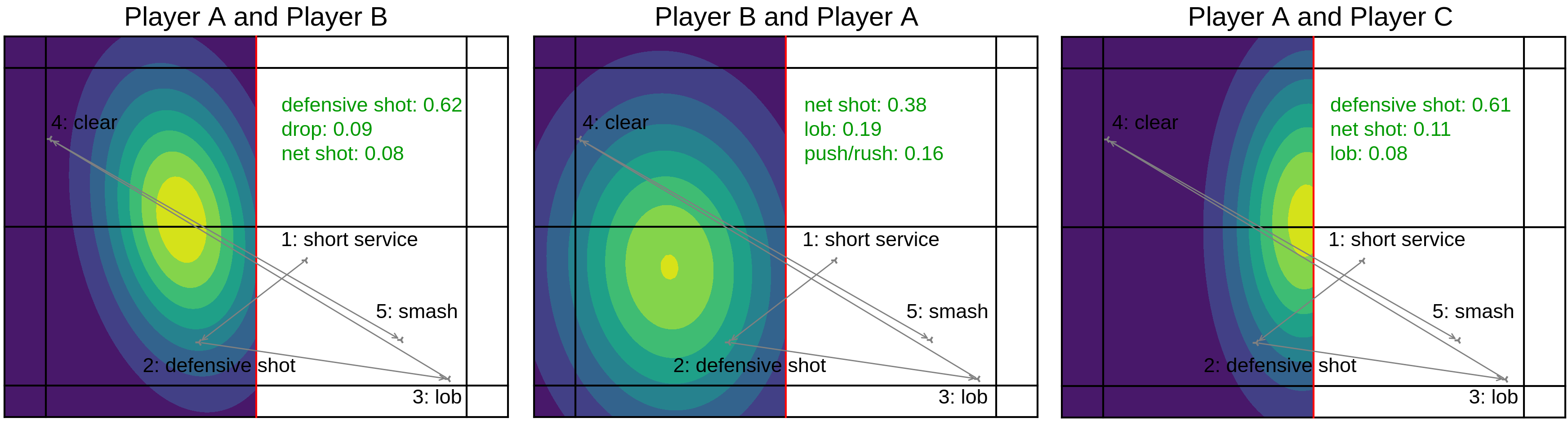}
  \caption{Illustration of three different matchups. The latter one of each case is the player returning the next stroke. That is, Player B is in the left figure, Player A is in the middle figure, and Player C is in the right figure.
  Black lines represent the court with the net in red.
  The top three shot types and corresponding probabilities are shown in green text.}
  \label{fig:case_study}
\end{figure*}

An extensive ablation study was conducted to verify the design of ShuttleNet. We developed six variants to investigate the relative contributions of different components introduced in ShuttleNet: 1) w/o L, which is ShuttleNet without the Transformer-based rally extractor, 2) w/o A, which is ShuttleNet without information of Player A in the Transformer-based player extractor, 3) w/o B, which is ShuttleNet without information of Player B in the Transformer-based player extractor, 4) w/o $\alpha$, which is ShuttleNet without information weights in the position-aware gated fusion network, 5) w/o $\beta$, which is ShuttleNet without position weights in the position-aware gated fusion network, and 6) w/o TAA, which is ShuttleNet with the type-area-attention mechanism replaced by the self-attention mechanism using concatenation of shot types and area.

It should be noted that the related terms of w/o L, w/o A, and w/o B were also removed in the fusion network and the fusion network was removed in w/o A + w/o B, w/o A + w/o L, and w/o B + w/o L since there is only one context information of each. Table \ref{tab:experiment-ablation} shows the results with $\tau=8$. We summarize the observations as follows.

\begin{table}
  \centering
  \begin{tabular}{c|ccc}
    \toprule
    Model & CE & MSE & MAE \\
    \midrule
    w/o L                     & 1.9917 & 1.6471 & 1.4085 \\
    w/o A                     & 1.9900 & 1.6418 & 1.4111 \\
    w/o B                     & 1.9848 & 1.6347 & 1.4071 \\
    w/o A + w/o B             & 2.4169 & 1.6525 & 1.4113 \\
    w/o A + w/o L             & 2.1693 & 2.7223 & 1.8647 \\
    w/o B + w/o L             & 2.0805 & 2.6625 & 1.8343 \\
    \midrule
    w/o $\beta$               & 2.1197 & 1.6344 & 1.4002 \\
    w/o $\alpha$              & 1.9822 & 1.6405 & 1.4088 \\
    \midrule
    w/o TAA                   & 1.9880 & 1.6628 & 1.4290 \\
    \midrule
    ShuttleNet (Ours)            & \textbf{1.9802} & \textbf{1.5856} & \textbf{1.3802} \\ 
    \bottomrule
\end{tabular}
  \caption{Ablation study of our model.}
  \label{tab:experiment-ablation}
\end{table}

\noindent\textbf{The effect of each context.} It is obvious that removing any one context in ShuttleNet results in a significant performance drop.
Also, as expected, using a single context leads to inferior performance in both shot types and area prediction.
The results verify the reasonable and effective design of our model.
Further, using only the rally context deteriorates the shot performance more substantially, while using only the context of either Player A or Player B negatively impacts more on area performance.
We suggest that the locations of the returns rely more on current progress, while the types of the returns can be more likely affected by the styles of each player. Meanwhile, our model demonstrates the ability of incorporating the use of each context.

\noindent\textbf{The performance of PGFN.} When we discard either fusing weights or position weights, the performance degrades in comparison with ShuttleNet. The results suggest that fusing the information of contexts is effective for different significance.
Also, position weights play an important role in the discrepancy of the importance of each stroke.

\noindent\textbf{Comparison with self-attention mechanism.} To testify the effective of the proposed type-area-attention mechanism, we compared it to the original self-attention mechanism. It is clear that applying the self-attention mechanism reduces the performance by 0.4\% on CE, 4.6\% on MSE, and 3.4\% on MAE in contrast to ShuttleNet.
These results signify that binding shot types and area on the same position is inadequate, whereas attending to different positions enhances the ability to capture the dependency from various aspects.

\subsection{Case Study: A Usage of Stroke Forecasting}
Analyzing returning strategies in different matchups with the same situation can help understand the possible strategies that the player may use by considering past information to formulate tactics.
In this case, the goal is to predict the shot type and area coordinates that the player returns at the sixth stroke of different matchups. Figure \ref{fig:case_study} shows three top-ranking players of men's singles. It shows that these players are likely to return with a passive shot type when encountering a smash from the opponent.
However, area distributions are quite different with respect to the players. The distribution of Player A is more in the middle of court, while both Player B and Player C are closer to the net.
Since the fourth stroke is at back court, if the return is nearer the net, the opponent will have a greater distance to move, which will consume more energy of the opponent. This case demonstrates a scenario analysis of stroke forecasting in badminton, and our model is capable of assisting not only coaches for tactic investigation but communities for storytelling.

\section{Conclusions and Future Works}
\label{sec:conclusion}

In this paper, we present ShuttleNet for tackling the challenging stroke forecasting problem.
Based on the encoder-decoder architecture, our model incorporates rally information and player information with two extractors.
In addition, a position-aware gated fusion network is proposed leveraging information dependency and position weights to decide the importance of rally contexts and contexts of the players for returning each strokes.
The quantitative evaluation conducted on the real-world dataset demonstrates the effectiveness of our proposed approach compared to state-of-the-art baselines.
For future work, we plan to extend our model to cope with extra conditions, \textit{e.g.}, win and loss, which can be analyzed for advanced tactic investigation.

\section{Acknowledgments}

This work was supported by the Ministry of Science and Technology of Taiwan under Grants MOST-110-2627-H-A49-001 and MOST-109-2221-E-009-114-MY3.

\bibliography{reference}

\end{document}